\pdfoutput=1

\documentclass[11pt]{article}

 \usepackage[]{acl}

\usepackage{times}
\usepackage{latexsym}

\usepackage[T1]{fontenc}

\usepackage[utf8]{inputenc}

\usepackage{microtype}
\usepackage[mathscr]{eucal}
\usepackage{amsmath}
\usepackage{graphicx}
\usepackage{color}

\title{Internal and External Knowledge Interactive Refinement Framework for Knowledge-Intensive Question Answering}

\author{Haowei Du,    Dongyan Zhao}

\begin{document}
\maketitle

\begin{abstract}
Recent works have attempted
to integrate external knowledge into LLMs to address the limitations and
potential factual errors in LLM-generated content. However, how to retrieve the correct knowledge from the large amount of external knowledge imposes a challenge.
To this end, we
empirically observe that LLMs have already encoded rich knowledge in their pretrained parameters and utilizing these internal knowledge improves the retrieval of external knowledge when applying them to knowledge-intensive
tasks. In this paper, we propose a new internal and external knowledge interactive refinement paradigm
dubbed IEKR to utilize internal knowledge in LLM to help retrieve relevant knowledge from the
external knowledge base, as well as exploit the external knowledge to refine the hallucination of generated internal knowledge. By simply adding a prompt
like “Tell me something about” to the LLMs, we try
to review related explicit knowledge and insert
them with the query into the retriever for external retrieval. The external knowledge is utilized to complement the internal knowledge into input of LLM for answers. We conduct experiments on 3 benchmark datasets in knowledge-intensive question answering task with different LLMs and domains, achieving the new state-of-the-art. Further analysis shows the effectiveness of different modules in our approach.
\end{abstract}

\section{Introduction}
Large Language Models (LLMs) have shown remarkable
abilities for human language processing
and extraordinary scalability and adaptability in
few- or zero-shot settings \cite{brown2020language,ouyang2022training,chowdhery2023palm}. In spite of LLM’s ability to
generate plausible-sounding text, hallucination can occur when the model produces text that includes facts or claims
that are fictional or misleading
rather than providing reliable and truthful information \cite{yao2022react, bang2023multitask}. 
To mitigate these limitations, recent works propose retrieval augmented generation (RAG) \cite{lewis2020retrieval, izacard2022few, khattab2022demonstrate} to integrates external
knowledge retrieved into the generative process. 
However, there is inevitably a gap between the input text
and the needed knowledge in retrieval \cite{ma2023query} and how to retrieve the right knowledge remains a challenge.
To this end, we empirically discover the internal knowledge encoded in LLM parameters help the retriever to obtain the correct external knowledge in demand for knowledge-intensive tasks.

\begin{figure}[h]
    \centering
\includegraphics[width=0.5\textwidth]{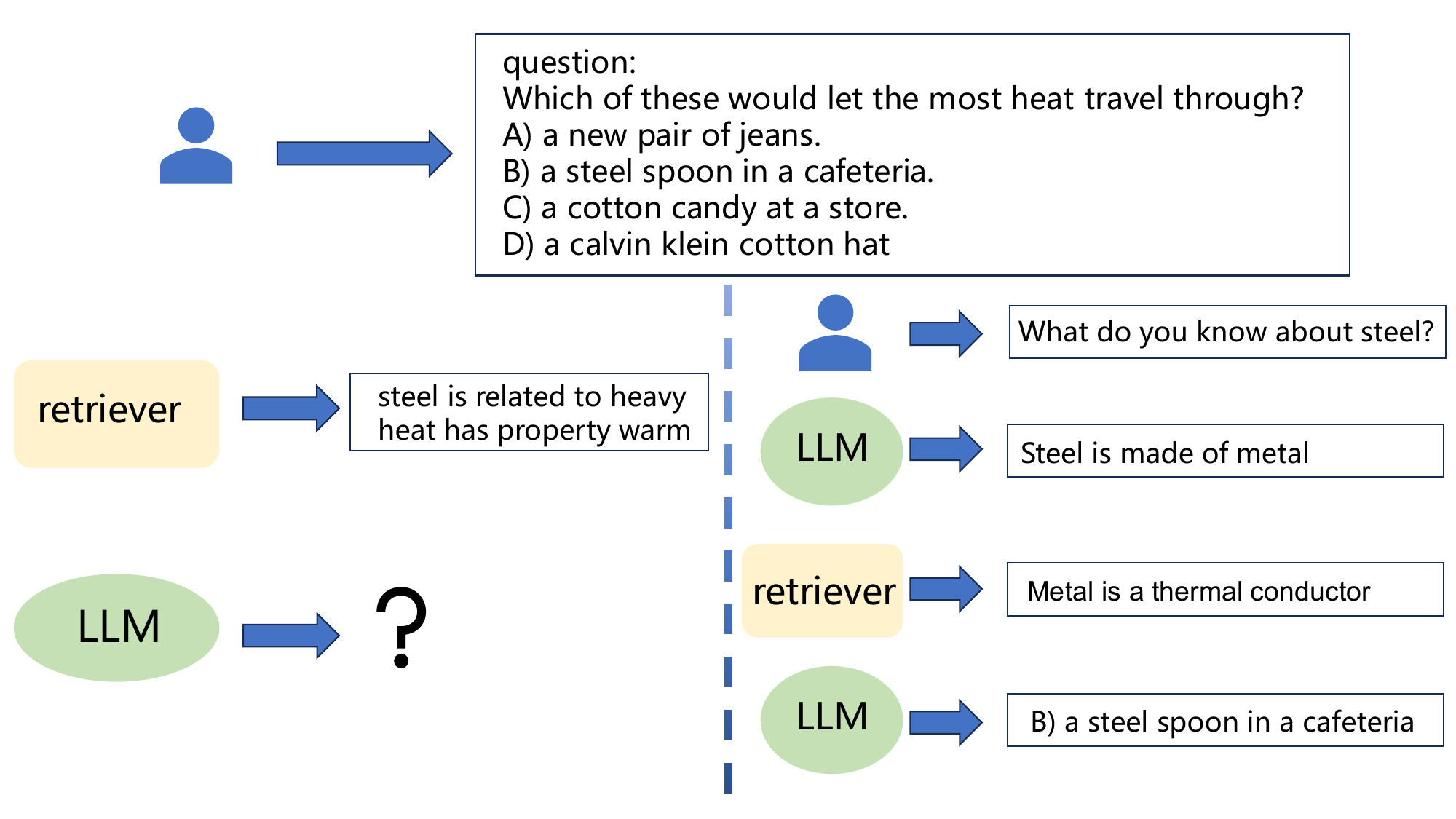}
    \caption{One example from OpenbookQA dataset}
    \label{case1}
\end{figure}

In Figure \ref{case1}, the query is ``Which of these would let the most heat travel through? A) a new pair of jeans. B) a steel spoon in a cafeteria. C) a cotton candy at a store. D) a calvin klein cotton hat'', and the needed knowledge to answer the question is ``Metal is a thermal conductor''. There is a gap between the option in query ``steel'' and this external knowledge, so the retriever chooses many relevant but not needed knowledge like ``steel is related to heavy, heat has property warm'' which distracts the LLM to answer the question. However, the knowledge gap can be filled by prompt the LLM to reflect on its internal knowledge about the steel.

To improve the retrieval of needed external knowledge for question answering (QA), we propose our internal and external knowledge interactively refinement framework (IEKR), where the internal knowledge within LLM is utilized to retrieve needed knowledge in external knowledge base (KB), and the external knowledge retrieved is incorporated into complementing the internal knowledge. Specifically, first we prompt the LLM to generate the intrinsic knowledge about the concepts in the query. Then we input the internal knowledge along with the query to the language model (LM) retriever and get the top-k knowledge sentences from the external KB. The internal and external knowledge is inputted to the reader to answer the question.
We conduct experiments on three benchmark knowledge-intensive QA datasets across different domains, OpenbookQA, CommonsenseQA and MedQA, becoming the new state-of-the-art. Further experimental results demonstrate the effectiveness of internal knowledge to retrieve the needed external knowledge, as well as the complement of internal knowledge with external knowledge. 

To conclude, we summarize the contributions of
this work as follows:
\textbf{1.} We are the first to incorporate internal knowledge within LLM to retrieve needed external knowledge for knowledge-intensive QA, filling the gap between the input text and the needed knowledge in retrieval.
\textbf{2.} We conduct experiments on three benchmark datasets with different LLMs across different domains, and derive the SOTA performance. Further experiments show the effectiveness of different modules of our approach.

\section{Related Work}

Previous
studies have shown that PLMs implicitly contain
a large amount of knowledge. \citet{petroni2019language}
have shown that such language models can be used
in a KB completion task by
converting KB relations into natural language templates. Based on this finding, researchers attempt to
treat the PLM as a knowledge base. Some studies \cite{bosselut2019comet, west2021symbolic} employ PLMs to construct knowledge graphs
automatically. Meanwhile, some others \cite{shwartz2020unsupervised, li2022eliciting} find that the knowledge
possessed by the PLMs can be used to enhance the
model’s performance in downstream tasks. To date,
several work \cite{wang2022pinto,zelikman2022star} attempt to utilize
PLMs to generate free-text rationales for reasoning. Our approach differs from previous works in
that we aim to utilize the internal knowledge in LLMs to enhance the external knowledge retrieval.


Using interactive question-knowledge
alignment, \citet{zhang2023mitigating} presents a method
for mitigating language model hallucination Their
proposed approach focuses on aligning generated
text with relevant factual knowledge, enabling
users to interactively guide the model’s responses
to produce more accurate and reliable information. This technique aims to improve the quality and factuality of language model outputs by
involving users in the alignment process. LLMAUGMENTER \cite{peng2023check} improves LLMs
using external knowledge and automated feedback.
It highlights the need to address the limitations and
potential factual errors in LLM-generated content.
This method involves incorporating external knowledge sources and automated feedback mechanisms
to enhance the accuracy and reliability of LLM
outputs. By doing so, the paper aims to mitigate
factual inaccuracies and improve the overall quality of LLM-generated text. Similarly, \citet{li2023chain} introduces a framework called “Chain of
Knowledge” for grounding LLMs with structured
knowledge bases. Grounding refers to the process
of connecting LLM-generated text with structured
knowledge to improve factual accuracy and reliability. This approach aims to improve the alignment of LLM-generated content with structured
knowledge, reducing the risk of generating inaccurate or hallucinated information. These methods neglect the internal knowledge within the LLM and there remains a gap between between the input text
and the needed knowledge in retrieval.

\section{Task Formulation}
We focus on the multi-choice QA task \cite{robinson2022leveraging}. The query includes a natural text question and several candidates and the model needs to choose an answer from the candidates:
\begin{equation}
\hat{\mathbf{a}}=\underset{\mathbf{a} \in \mathcal{C}}{\arg \max } P(\mathbf{a} \mid \mathbf{q})
\end{equation} where $\mathcal{C}$ denotes the answer candidates and $q$ denotes the question.
There is an external KB to provide external knowledge for the model. The KB contains a series of factual triples and each triple (fact) is composed of two entities and one relation. A KB can be denoted as $\mathcal{G} = \{(e,r,e')| e,e' \in E, r\in R\}$, where $G$ denotes the KB, $E$ denotes the entity set and $R$ denotes the relation set.

\section{Methodology}

\begin{figure*}[h]
    \centering
\includegraphics[width=0.95\textwidth]{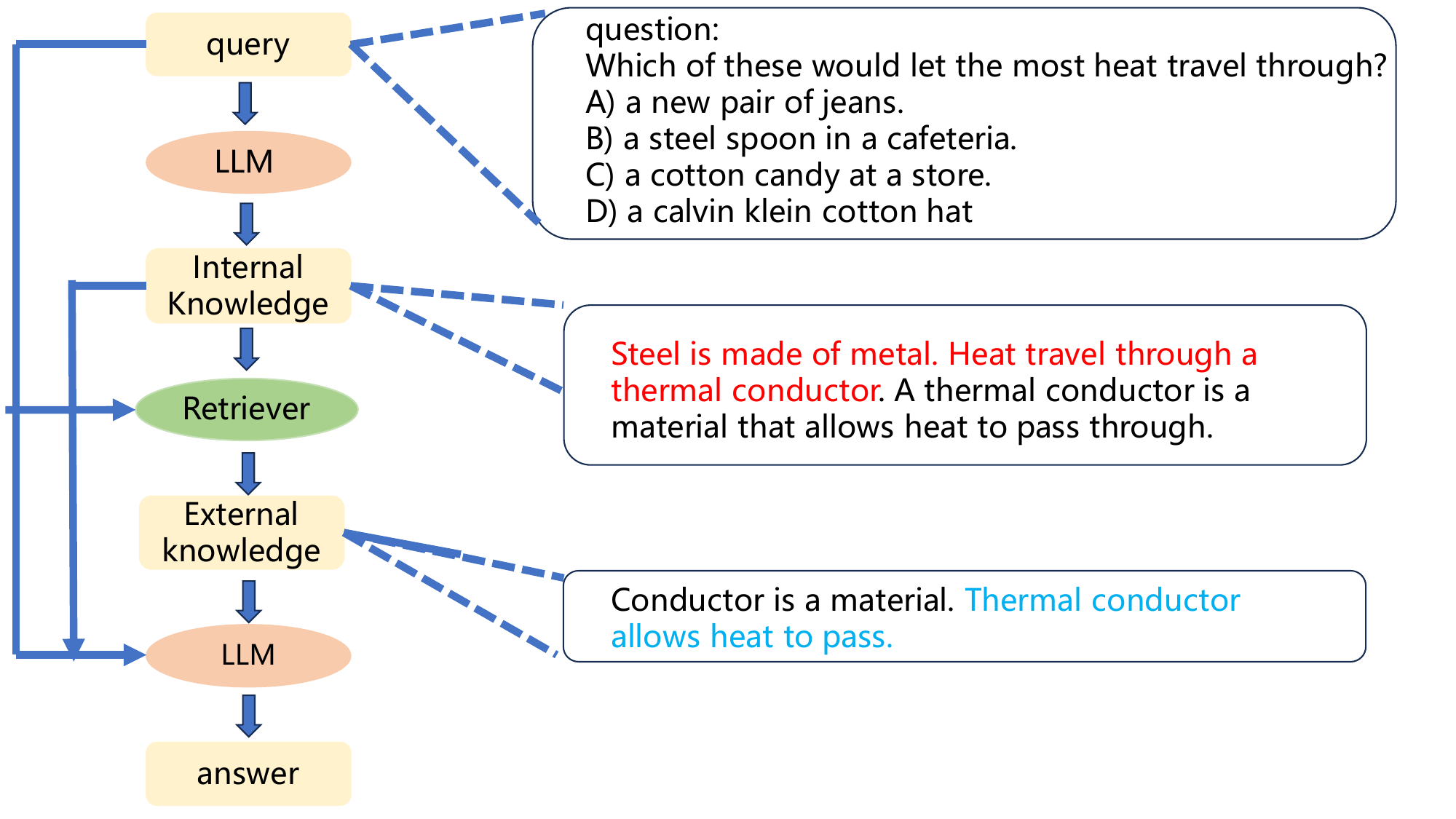}
    \caption{The pipeline of our approach. Our model composes 2 modules: a retriever $\mathcal{R}$ to retrieve the relevant knowledge facts from the external KB;  an LLM reader $\mathcal{A}$ to answer the query based on the internal knowledge within the model as well as retrieved external knowledge. Red denotes internal knowledge needed for the question and blue denotes the external knowledge retrieved with internal knowledge.}
    \label{pipe}
\end{figure*}

In this part, we introduce the architecture of our approach. Our model composes 2 modules: an LM retriever $\mathcal{R}$ is used to retrieve the relevant knowledge facts from the external KB $\mathcal{G}$;  an LLM reader $\mathcal{A}$ is utilized to answer the query based on the internal knowledge in the model as well as retrieved external knowledge.

For an input query, first we prompt the LLM $\mathcal{A}$ to reflect on its internal knowledge about the query entities. Secondly we utilize the internal knowledge as well as query to retrieve the relevant external knowledge by retriever $\mathcal{R}$ to complement internal knowledge. Finally the internal knowledge, the complementary external knowledge as well as the query are inputted into LLM $\mathcal{A}$ to derive the answer. 

\subsection{Internal Knowledge Reflection}
In this part, we aim to dig into the LLM about the internal knowledge about the entities in query.
We utilize off-shelf Named Entity Recognition model to extract the named entities in the query. For example, as to  ``Frilled sharks and angler fish live far beneath the surface of the ocean'', we extract the ``Frilled sharks'', ``angler fish'' and ``ocean'' as the named entities. 

For each entity, we construct the prompt like ``tell me something about [entity]'' to ask LLM $\mathcal{A}$ to generate the internal knowledge about it. We arrange the generation about each entity sequentially as the internal knowledge of LLM $\mathcal{A}$ of the current query:

\begin{align}
    \mathbf{k_i} &= \mathcal{A}(\mathbf{P} + \mathbf{e_i}) \\
   \mathbf{IK} &= \mathbf{Concate}[\mathbf{k_1} || \mathbf{k_2} || \cdots || \mathbf{k_n}] 
\end{align}
where $\mathbf{P}$ denotes the reflection prompt to LLM $\mathcal{A}$, $n$ denotes the number of entities, $\mathbf{e_i}$ denotes the i-th entity in query.

\subsection{External Knowledge Retrieval}

In this part, we utilize a pretrained LM to retrieve the relevant knowledge from the external KB $\mathcal{G}$ base on the internal knowledge $IK$ and query as the complement. The KB contains a large set of triples of the form (h, r, t), like (ice, HasProperty, cold), where h and t represent head and tail entities in the entity set $V$ and r is a certain relation
type from the pre-defined set. To reduce the search space, we following \cite{zhang2022greaselm} to perform entity linking to $\mathcal{G}$ to derive an initial set of nodes $\mathbf{V_{linked}}$ and prune the original KB $\mathcal{G}$ to keep the entities within 2 hops from $\mathbf{V_{linked}}$. 
We convert the triples into natural language templates, such as ``(ice, HasProperty, cold)'' is converted into ``Ice has the property of cold''.

The retriever contains a pretrained cross-encoder with transformer architecture, on top of which there is a classifer to predict the similarity score between the internal knowledge and the external knowledge sentence.
We choose the top-k external sentences based on the similarity score as complements of internal knowledge:
\begin{align}
    \mathbf{s_i} &= \mathbf{MLP}(\mathcal{R}( \left[ \mathbf{q}:\mathbf{IK}:\mathbf{K_i} \right])) \\
   \mathbf{EK} &= \mathbf{Concate}[\mathbf{K_{(1)}} || \mathbf{K_{(2)}} || \cdots || \mathbf{K_{(m)}}] 
\end{align}
where $\mathbf{q}$ denotes the query, $\mathbf{K_i}$ denotes the i-th external sentence, $\mathbf{K_{(i)}}$ denotes the i-th external sentence after sorting by $\mathbf{s_i}$ and $m$ denotes the number of external knowledge sentence.

\subsection{Answer Generation}
In this part, we input the internal knowledge, external knowledge as well as the query into the LLM $\mathcal{A}$ to generate the answer. The cross-entropy loss is utilized to optimize the model:

\begin{align}
    L  &= -\mathbf{log} \, p(a | q, \mathbf{IK^R}, \mathbf{EK}) \\
    &= -\mathbf{log} \prod_{i=1}^l p(a_i |  q, \mathbf{IK^R}, \mathbf{EK}, a_{<i}) \\
    & = -\sum_{i=1}^l \mathbf{log} \, p(a_i | q, \mathbf{IK^R}, \mathbf{EK}, a_{<i})  
\end{align}
where $a$ denotes the answer generated and $l$ denotes the answer length.

\section{Experiments}

\subsection{Datasets}
We evaluate IEKR on three diverse multiple-choice question answering datasets across two
domains: CommonsenseQA \cite{talmor2018commonsenseqa} and OpenBookQA \cite{mihaylov2018can} as commonsense
reasoning benchmarks, and MedQA-USMLE \cite{jin2021disease} as a clinical QA task.
CommonsenseQA is a 5-way multiple-choice question answering dataset of 12,102 questions that
require background commonsense knowledge beyond surface language understanding. We perform
our experiments using the in-house data split of \cite{lin2019kagnet} to compare to baseline methods.
OpenbookQA is a 4-way multiple-choice question answering dataset that tests elementary scientific
knowledge. It contains 5,957 questions along with an open book of scientific facts. We use the
official data splits from \cite{mihaylov2018can}.
MedQA-USMLE is a 4-way multiple-choice question answering dataset, which requires biomedical
and clinical knowledge. The questions are originally from practice tests for the United States
Medical License Exams (USMLE). The dataset contains 12,723 questions. We use the original data
splits from \cite{jin2021disease}.

\begin{table}
\centering
\begin{tabular}{lcc}
\hline Methods & Dev  & IHtest\\
\hline 
MHGRN  \cite{feng2020scalable}& 74.5 & 71.1  \\
QA-GNN \cite{yasunaga2021qa} & 76.5 & 73.4 \\ 
GREASELM  \cite{zhang2022greaselm} & 78.5  & 74.2\\
RumiDeBERTa \cite{yao2023knowledge} & - & 74.3 \\
GrapeQA \cite{taunk2023grapeqa} & - & 74.9 \\ 
Dragon  \cite{yasunaga2022deep} & -  & 76.0\\
\hline
IEKR & \textbf{87.9} &  \textbf{93.7} \\
\hline
\end{tabular}
\caption{Results on CommonsenseQA dataset compared with LM+GNN methods. As the official test is hidden, here we report the in-house Test (IHtest) accuracy,
following the data split of Lin et al. (2019).}
\label{result-com-non}
\end{table}

\begin{table}
\centering
\begin{tabular}{lcc}
\hline Methods & Dev & Para\\
\hline 
GPT-3 \cite{xu2021human} & 73.0 & 175B \\
UnifiedQA \cite{khashabi2020unifiedqa}& 79.1 & 11B \\
Flan-T5 \cite{chung2022scaling} & 83.2 & 3B \\
RumiFlanT5 \cite{yao2023knowledge}& 84.3 & 3B \\
GKP \cite{liu2021generated} & 85.3 & 11B \\
\hline
IEKR &  \textbf{87.9} & 3B \\
\hline
\end{tabular}
\caption{Results on CommonsenseQA dataset compared with LLM based methods. We adopt accuracy as the metric to evaluate the performance. GKP denotes Generated
Knowledge Prompting. $para$ denotes the parameter number in the model.}
\label{result-com}
\end{table}

\subsection{Implementation Details}

We utilize the widespread Flan-T5 (3B) \cite{chung2022scaling} as the LLM $\mathcal{A}$ for CommonsenseQA and OpenbookQA. Because MedQA-USMLE requires more domain-specific knowledge, we choose the LLama2 (7B) \cite{touvron2023llama} as the LLM $\mathcal{A}$. We adopts the same verifier $\mathcal{V}$ for the three datasets, which is initialzed with LLama2 (7B).
To reduce computation cost and keep prior knowledge in LLM, we use LoRA \cite{hu2021lora}, which freezes the pretrained model weights and injects trainable rank decomposition matrices into each
layer of the LLM $\mathcal{A}$. So the number of trainable parameters
of IEKR-7B is 4.5M, only 0.06\% of total
parameters of LLaMA2-7B.

We use ConceptNet \cite{speer2017conceptnet}, a general-domain knowledge graph, as our external knowledge source $\mathcal{G}$ for both CommonsenseQA and OpenbookQA. It has 799,273 nodes and 2,487,810 edges in total. For MedQA-USMLE, we use a self-constructed knowledge
graph that integrates the Disease Database portion of the Unified Medical Language System \cite{bodenreider2004unified} and DrugBank \cite{wishart2018drugbank}. The knowledge graph contains 9,958
nodes and 44,561 edges. 

We adopt the pretrained dense retriever BGE-Reranker \cite{chen2024bge} to initialize the retriever $\mathcal{R}$. The number of cases $N$ sampled from training dataset to finetune $\mathcal{R}$ is set to 500. The number of external knowledge sentences $m$ retrieved by $\mathcal{R}$ wth the query and internal knowledge is set to 50. 
We utilize AdamW as the optimizer to train the LLM $\mathcal{A}$ and the learning rate is set to 3e-5. Training batch size is set to 2.

\begin{table}
\centering
\begin{tabular}{lcc}
\hline Methods & Test\\
\hline
Dragon \cite{yasunaga2022deep} & 72.0 \\
RumiDeBERTa \cite{yao2023knowledge} & 76.0 \\
ALBERT+KG \cite{lan2019albert} & 81.0  \\
HGN \cite{yan2020learning}  & 81.4 \\
AMR-SG \cite{xu2021dynamic} & 81.6 \\
ALBERT+KPG \cite{wang2020connecting} &  81.8 \\
DEKCOR \cite{xu2021fusing} & 82.2 \\
QA-GNN  \cite{yasunaga2021qa} & 82.8 \\
GREASELM \cite{zhang2022greaselm} &  84.8 \\
\hline
IEKR & \textbf{92.1}  \\
\hline
\end{tabular}
\caption{Results on OpenbookQA dataset compared with LM+GNN based methods. }
\label{result-open-non}
\end{table}

\subsection{Baselines}
We compare our methods with 2 groups of baselines: LM based methods with graph neural network (GNN) and LLM based methods.
LM+GNN methods utilize GNN to incorporate the external knowledge from KB for knowledge-intensive QA. Because GNN involves much computation cost which does not apply to LLM, LLM based methods adopt the external knowledge text in end-to-end training. 

For LM+GNN methods, we compare our method with 
RoBERTa-Large 
RGCN \cite{schlichtkrull2018modeling},
GconAttn \cite{wang2019improving}, 
KagNet \cite{lin2019kagnet},
RN \cite{santoro2017simple},
MHGRN  \cite{feng2020scalable},
QA-GNN \cite{yasunaga2021qa}, 
GREASELM  \cite{zhang2022greaselm},
RumiDeBERTa \cite{yao2023knowledge}, 
GrapeQA \cite{taunk2023grapeqa}, and
Dragon  \cite{yasunaga2022deep} for OpenbookQA and CommonsenseQA; as well as
SapBERT \cite{liu2020self},
QA-GNN \cite{yasunaga2021qa},
GREASELM  \cite{zhang2022greaselm}, and
GrapeQA \cite{taunk2023grapeqa} for MedQA.

For LLM based methods, we compare our method with 
GPT-3 \cite{xu2021human},
UnifiedQA \cite{khashabi2020unifiedqa},
Flan-T5 \cite{chung2022scaling},
RumiFlanT5 \cite{yao2023knowledge},
DeBERTa-xxlarge \cite{xu2021human},
GKP \cite{liu2021generated}, and GenMC \cite{huang2022clues} for OpenbookQA and CommonsenseQA; as well as GPT-Neo \cite{black2022gpt},
LLama2 \cite{touvron2023llama}, and 
RumiLLama2 \cite{yao2023knowledge} for MedQA.
For RAG based LLM, we compare with FLARE\cite{jiang2023active} and ReFeed \cite{yu2023improving}, and replace the original backbone from GPT-3.5 to the same as ours.

\begin{table}
\centering
\begin{tabular}{lcc}
\hline Methods & Test & Para\\
\hline 
GPT-3 few shot \cite{xu2021human} & 73.0 & 175B \\
T5 \cite{raffel2020exploring} &  83.2  & 3B \\
T5+KB \cite{pirtoacaai2} &  85.4  & 11B \\
Flan-T5 \cite{chung2022scaling} & 86.5 & 3B \\
ReFeed \cite{yu2023improving} & 87.1 & 3B \\
UnifiedQA \cite{khashabi2020unifiedqa} & 87.2 & 11B \\
RumiFlanT5 \cite{yao2023knowledge} & 87.3 & 3B \\
FLARE \cite{jiang2023active}& 88.6 & 3B \\
GenMC \cite{huang2022clues}& 89.8 & 11B \\
\hline
IEKR &  \textbf{92.1} & 3B \\
\hline
\end{tabular}
\caption{Results on OpenbookQA dataset compared with LLM based methods. `Para'' denotes the parameter number in the model. RumiFlanT5 are trained by using FlanT5-3B to replace RumiDeBERTa \cite{yao2023knowledge} as the backbone for fair comparison.}
\label{result-open}
\end{table}

\subsection{Results}

Our results in Tables \ref{result-com} and \ref{result-com-non} demonstrate a consistent improvement on the CommonsenseQA dataset compared with existing LM+GNN based methods and LLM-based methods, becoming the new state-of-the-art. Compared with competitive LM+GNN methods, GREASELM and Dragon, our model outperforms by 9.4 accuracy on dev set and 17.7 accuracy on in-house test set. Compared with best LLM based method, Generated
Knowledge Prompting, our model outperforms by 2.6 accuracy on dev set.

In Tables \ref{result-open-non} and \ref{result-open}, our improvements significantly outperforms the existing LM+GNN and LLM based methods on OpenbookQA dataset, becoming the new SOTA. IEKR improves the performance by 7.3 accuracy compared with the best LM+GNN method GREASELM, and 2.3 accuracy compared with the best LLM method GenMC on test set.
It demonstrates the effectiveness of our internal and external knowledge interactively refinement framework. 

Our reported results thus
far demonstrate the viability of our method in
the general commonsense reasoning domain. Further, we explore whether IEKR could be adapted to other domains by evaluating
on the MedQA-USMLE dataset.
Our results in Tables \ref{result-med-non} and \ref{result-med} demonstrate that IEKR outperforms
SOTA LM+GNN method GrapeQA by 11.4 accuracy and LLM based method RumiLLama2 by 3.5 accuracy. It demonstrates that with different LLMs, our approach shows stable improvement over the knowledge intensive QA task across different domains. Our method effectively reflects on useful internal knowledge within the model, and utilize it to enhance the retrieval of external knowledge for QA task.

\section{Analysis}

In this part, we conduct ablation studies to evaluate different modules of our approach. Then we generalize our method to other QA tasks and compare our methods with different RAG based methods with the same backbone. Finally, we conduct experiments with different numbers of external knowledge sentences $m$.

\subsection{Ablation Study}
In our approach, there are 2 modules: Internal Knowledge Reflection, External Knowledge Retrieval. We successively evaluate the importance of different modules by removing the respect module.

\paragraph{Does internal knowledge reflection matter?}
In this ablation, we remove the process of digging into the LLM about the
internal knowledge about the entities in query. We directly utilize the query to retrieve relevant external knowledge as inputs into LLM $\mathcal{A}$ along with the query to derive the answer.
In Table \ref{ablation}, when removing the internal knowledge reflection, our model drops by 2.3 accuracy on CommonsenseQA, 3.4 accuracy on OpenbookQA, and 3.3 accuracy on MedQA. It demonstrates that without the internal knowledge within LLM, the retriever struggles to retrieve enough needed knowledge from external KB only with the query. However, compared with direct finetuning, this ablation model outperforms by 1.1, 2.2, and 0.9 accuracy on CommonsenseQA, OpenbookQA and MedQA, which shows our retriever still derives some useful external knowledge from KB $\mathcal{G}$ for the QA task.

\begin{table}
\centering
\begin{tabular}{lcc}
\hline Methods & Acc.  \\
\hline
SapBERT \cite{liu2020self} & 37.2 \\
QA-GNN \cite{yasunaga2021qa}& 38.0 \\
GREASELM  \cite{zhang2022greaselm} & 38.5 \\
GrapeQA  \cite{taunk2023grapeqa} & 39.5 \\
\hline
IEKR & 50.9 \\
\hline
\end{tabular}
\caption{Results on MedQA dataset compared with LM+GNN based methods.}
\label{result-med-non}
\end{table}

\begin{table}
\centering
\begin{tabular}{lcc}
\hline Methods & Acc. & Params \\
\hline
GPT-Neo \cite{black2022gpt} & 33.3 & 2.7B\\
LLama2 \cite{touvron2023llama} &  46.7 & 7B \\
RumiLLama2 \cite{yao2023knowledge} & 47.4 & 7B \\
\hline
IEKR & 50.9 & 7B \\
\hline
\end{tabular}
\caption{Results on MedQA dataset compared with LLM based methods. ``Params'' denotes the parameter number in the model. RumiLLama2 are trained by using LLama2-7B to replace RumiDeBERTa \cite{yao2023knowledge} as the backbone for fair comparison. We conduct experiments with 5 different random seeds and the p-value is less than 0.001.}
\label{result-med}
\end{table}

\begin{table}
\centering
\begin{tabular}{lccc}
\hline Methods & CSQA & OBQA & MedQA \\
\hline
Backbone & 90.3 & 86.5 &  46.7 \\
- internal & 91.4 & 88.7 & 47.6\\
- external & 92.2 & 89.8 & 48.2 \\
\hline
IEKR  & \textbf{93.7} & \textbf{92.1} & \textbf{50.9} \\
\hline
\end{tabular}
\caption{Ablation results on CommonsenseQA IHtest set, OpenbookQA test set, and MedQA test set. ``CSQA'' denotes CommonsenseQA, ``OBQA'' denotes OpenbookQA. ``-internal'' denotes removing the internal knowledge relection; ``- external'' denotes removing external knowledge retrieval; ``Backbone'' denotes directly finetuning the backbone model to generate the answer, i.e., FlanT5-3B for CSQA and OBQA.}
\label{ablation}
\end{table}

\begin{table}
\centering
\begin{tabular}{lccc}
\hline Number & CSQA & OBQA & MedQA \\
\hline
10 & 92.9 & 91.0 &  49.5 \\
30 & 93.2 & 91.7 & 50.3\\
50 & \textbf{93.7} & 92.1 & \textbf{50.9} \\
100 & 93.1 & \textbf{92.2} & 50.4 \\
\hline
\end{tabular}
\caption{Results with different number of external knowledge sentences. We use accuracy on test set as evaluation.}
\label{ext-num}
\end{table}

\paragraph{Does external knowledge retrieval matter?}
In this ablation, we remove the process of retrieving knowledge from external KB and alleviating hallucination with verifier. We prompt the model to derive the internal knowledge about the query entity and input the internal knowledge with query to the LLM for the answer.
In Table \ref{ablation}, when removing the external knowledge retrieval, our model drops by 1.5 accuracy on CommonsenseQA, 2.3 accuracy on OpenbookQA, and 2.7 accuracy on MedQA. It demonstrates that the knowledge derived from internal reflection does not contain the enough information to answer the question. The model still needs to retrieve from external KB to complement the internal knowledge for knowledge-intensive QA. However, compared with direct finetuning, this ablation model outperforms by 1.9, 3.3, and 1.5 accuracy on CommonsenseQA, OpenbookQA and MedQA, which demonstrates the effectiveness of prompting LLM to reflect on internal knowledge.



\begin{table*}[h]\small
\centering
\begin{tabular}{|c|c|}
\hline Dataset & Example \\
\hline CommonsenseQA & \begin{tabular}{l} 
A weasel has a thin body and short legs to easier burrow after prey in a what? \\
(A) tree (B) mulberry bush (C) chicken coop (D) viking ship \textcolor[rgb]{1,0,0}{(E) rabbit warren}
\end{tabular} \\
\hline OpenbookQA & \begin{tabular}{l} 
Which of these would let the most heat travel through? \\
\begin{tabular}{ll} 
(A) a new pair of jeans & \textcolor[rgb]{1,0,0}{(B) a steel spoon in a cafeteria} \\
(C) a cotton candy at a store & (D) a calvin klein cotton hat
\end{tabular}
\end{tabular} \\
\hline MedQA-USMLE & \begin{tabular}{l} 
A 57 -year-old man presents to his primary care physician with a 2-month \\
history of right upper and lower extremity weakness. He noticed the weakness \\
when he started falling far more frequently while running errands. Since then, \\
he has had increasing difficulty with walking and lifting objects. His past \\
medical history is significant only for well-controlled hypertension, but he says \\
that some members of his family have had musculoskeletal problems. His right \\
upper extremity shows forearm atrophy and depressed reflexes while his right \\
lower extremity is hypertonic with a positive Babinski sign. Which of the \\
following is most likely associated with the cause of this patients symptoms? \\
\begin{tabular}{ll} 
(A) HLA-B8 haplotype & (B) HLA-DR2 haplotype \\
\textcolor[rgb]{1,0,0}{(C) Mutation in SOD1} & (D) Mutation in SMN1
\end{tabular}
\end{tabular} \\
\hline
\end{tabular}
\caption{Examples of the Knowledge-intensive QA task for each of the datasets evaluated in this work. }
\end{table*}

\begin{table}
\centering
\begin{tabular}{lccc}
\hline Model & CSQA & OBQA & MedQA \\
\hline
Backbone & 90.3	& 86.5	& 46.7 \\
ReFeed	& 90.8	& 87.1	& 47.1\\
FLARE	& 91.5	& 88.6	& 48.2 \\
Ours	& 93.7	& 92.1	& 50.9 \\
\hline
\end{tabular}
\caption{Comparing with different RAG based methods with the same backbone as ours. We use accuracy on test set as evaluation.}
\label{different-RAG}
\end{table}

\subsection{Comparing with RAG methods}
In this part, we compare our method with existing competitive RAG based methods ReFeed \cite{yu2023improving} and FLARE \cite{jiang2023active} with the same backbone.

In Table \ref{different-RAG}, it shows our method significantly outperforms the two competitive RAG baselines by over 2.2 accuracy on different datasets. Our method does not need the LLM to have the ability to give an initial answer or ask follow-up query. We focus on the concrete factual knowledge within LLM, which provides more new and valuable information for external retrieval. Compared with FLARE, the follow-up queries generated by smaller LLM like LLama other than ChatGPT do not provide as much valuable information as our method for external retrieval. Moreover, our method does not need multi-time retrieval and only retrieves once based on the concrete internal knowledge within LLM but derives significant improvement. Considering the large size of external KB, We reduce the computation cost and improve the performance by utilizing internal knowledge for external retrieval in one time. Compared with ReFeed, the concrete internal knowledge in LLM provides more valuable information for external retrieval than the initial answer of LLM. It demonstrates the significant effectiveness and efficiency of our method over other RAG methods.
We highlighted the concrete factual knowledge within LLM instead of the initial answer or follow-up query of LLM in FLARE and ReFeed. Considering the setting of multi-choice QA, the choices has been included in query and initial answer of LLM will not provide much new knowledge beyond the content of query. However, our method prompts the LLM to provide concrete internal knowledge about the linguistic component of query. For example, we ask the LLM "tell me something about heat travel'' and the LLM provides the internal knowledge "Heat travel through a thermal conductor'' instead of "the answer is B'' in ReFeed. To our best knowledge, we are the first to introduce concrete internal knowledge within LLM for external retrieval.

On the other hand, we design the module of verifier shared by different datasets to modify the internal knowledge based on the retrieved external knowledge and combine the two sources of knowledge to answer the question. In FLARE and ReFeed, the internal knowledge contains the initial answer or follow-up query and may introduce distracting contents especially for smaller LLM like Llama-7b.

Retrieval from external KB can bring much computation cost because of the vast amount of knowledge. Our method does not need multi-time retrieval like FLARE and only retrieves once based on the concrete internal knowledge within LLM to derive significant improvement.

\subsection{Generalization to Other QA Tasks}
In this section, we apply our methods to the widely-recognized open-domain question answering (QA) dataset, 2WikiMultihopQA \cite{ho2020constructing}. Our approach follows the framework established by FLARE, utilizing the Llama-7b model as the foundational backbone, to showcase the generalization capabilities of our methodology. Unlike multiple-choice question answering (MCQA), where the objective is to select the correct answer from a given set of options, open-domain QA requires generating free-form text responses to queries. To evaluate the performance of our method, we employ Exact Match (EM) and F1 score as the primary metrics.

As illustrated in Table \ref{open-domain}, our method surpasses the performance of FLARE, achieving an improvement of 1.4 in EM and 1.9 in F1 score. Notably, while FLARE necessitates multiple retrievals of external knowledge to formulate an answer, our method accomplishes this with a single retrieval step. This significant difference underscores the efficiency and effectiveness of our approach when applied to the open-domain QA dataset. The ability to reduce retrieval frequency without compromising accuracy not only highlights the robustness of our method but also suggests potential for enhanced scalability and practicality in real-world applications.

\subsection{Different Retrieval Number }
In this section, we examine the impact of varying the number of external knowledge sentences, denoted as $m$, on our method's performance. For our primary experiments, we set $m$ to 50, and we conducted additional experiments varying $m$ from 10 to 100. As shown in Table \ref{ext-num}, we observed that generally, retrieving a greater number of knowledge sentences from the external knowledge base (KB) enhances the performance of our method. The inclusion of more external knowledge provides the model with a richer set of factual information pertinent to the query, thereby improving the verifier's ability to mitigate hallucinations when revising internal knowledge.

However, an interesting phenomenon occurs when $m$ is set to 100: the Integrated External Knowledge Retriever (IEKR) exhibits a slight decline in performance. This drop can be attributed to the longer external knowledge context, which includes some relevant but unnecessary sentences. These extraneous sentences can distract the model, complicating its reasoning process and ultimately impairing its ability to derive accurate answers. Therefore, while increasing the amount of retrieved external knowledge generally benefits performance, there is a threshold beyond which the inclusion of superfluous information can become counterproductive. This finding underscores the importance of optimizing the balance between the quantity and relevance of external knowledge in enhancing the model's reasoning capabilities.

\begin{table}
\centering
\begin{tabular}{lcc}
\hline Model & EM	& F1 \\
\hline
Backbone & 	60.1 & 	65.3 \\
ReFeed & 61.5 & 66.2 \\
FLARE &	62.0	& 66.7 \\
Ours & 	63.4	& 68.6 \\
\hline
\end{tabular}
\caption{Evaluation on 2WikiMultihopQA dataset with Llama-7b as the backbone. We utilize Exact match and F1 as the evaluation metrics.}
\label{open-domain}
\end{table}


\section{Conclusion}
In this work, we propose the internal and external knowledge interactively refinement framework, where the internal knowledge within LLM are utilized to retrieve needed knowledge in external KB, and the external knowledge retrieved are incorporated into revising the internal knowledge. We demonstrate our effectiveness on 3 benchmark datasets in knowledge-intensive QA with different LLMs across different domains.

\section*{Limitations}
 We propose the internal and external knowledge interactively refinement framework, and demonstrate our effiveness on 3 benchmark datasets in knowledge-intensive QA with different LLMs across different domains. The sizes of LLMs we use range from 3B to 7B, and we will conduct experiments with LLM larger than 7B in future research.



\bibliography{anthology,custom}

\begin{thebibliography}{55}
\expandafter\ifx\csname natexlab\endcsname\relax\def\natexlab#1{#1}\fi

\bibitem[{Bang et~al.(2023)Bang, Cahyawijaya, Lee, Dai, Su, Wilie, Lovenia, Ji, Yu, Chung et~al.}]{bang2023multitask}
Yejin Bang, Samuel Cahyawijaya, Nayeon Lee, Wenliang Dai, Dan Su, Bryan Wilie, Holy Lovenia, Ziwei Ji, Tiezheng Yu, Willy Chung, et~al. 2023.
\newblock A multitask, multilingual, multimodal evaluation of chatgpt on reasoning, hallucination, and interactivity.
\newblock \emph{arXiv preprint arXiv:2302.04023}.

\bibitem[{Black et~al.(2022)Black, Biderman, Hallahan, Anthony, Gao, Golding, He, Leahy, McDonell, Phang et~al.}]{black2022gpt}
Sid Black, Stella Biderman, Eric Hallahan, Quentin Anthony, Leo Gao, Laurence Golding, Horace He, Connor Leahy, Kyle McDonell, Jason Phang, et~al. 2022.
\newblock Gpt-neox-20b: An open-source autoregressive language model.
\newblock \emph{arXiv preprint arXiv:2204.06745}.

\bibitem[{Bodenreider(2004)}]{bodenreider2004unified}
Olivier Bodenreider. 2004.
\newblock The unified medical language system (umls): integrating biomedical terminology.
\newblock \emph{Nucleic acids research}, 32(suppl\_1):D267--D270.

\bibitem[{Bosselut et~al.(2019)Bosselut, Rashkin, Sap, Malaviya, Celikyilmaz, and Choi}]{bosselut2019comet}
Antoine Bosselut, Hannah Rashkin, Maarten Sap, Chaitanya Malaviya, Asli Celikyilmaz, and Yejin Choi. 2019.
\newblock Comet: Commonsense transformers for automatic knowledge graph construction.
\newblock \emph{arXiv preprint arXiv:1906.05317}.

\bibitem[{Brown et~al.(2020)Brown, Mann, Ryder, Subbiah, Kaplan, Dhariwal, Neelakantan, Shyam, Sastry, Askell et~al.}]{brown2020language}
Tom Brown, Benjamin Mann, Nick Ryder, Melanie Subbiah, Jared~D Kaplan, Prafulla Dhariwal, Arvind Neelakantan, Pranav Shyam, Girish Sastry, Amanda Askell, et~al. 2020.
\newblock Language models are few-shot learners.
\newblock \emph{Advances in neural information processing systems}, 33:1877--1901.

\bibitem[{Chen et~al.(2024)Chen, Xiao, Zhang, Luo, Lian, and Liu}]{chen2024bge}
Jianlv Chen, Shitao Xiao, Peitian Zhang, Kun Luo, Defu Lian, and Zheng Liu. 2024.
\newblock Bge m3-embedding: Multi-lingual, multi-functionality, multi-granularity text embeddings through self-knowledge distillation.
\newblock \emph{arXiv preprint arXiv:2402.03216}.

\bibitem[{Chowdhery et~al.(2023)Chowdhery, Narang, Devlin, Bosma, Mishra, Roberts, Barham, Chung, Sutton, Gehrmann et~al.}]{chowdhery2023palm}
Aakanksha Chowdhery, Sharan Narang, Jacob Devlin, Maarten Bosma, Gaurav Mishra, Adam Roberts, Paul Barham, Hyung~Won Chung, Charles Sutton, Sebastian Gehrmann, et~al. 2023.
\newblock Palm: Scaling language modeling with pathways.
\newblock \emph{Journal of Machine Learning Research}, 24(240):1--113.

\bibitem[{Chung et~al.(2022)Chung, Hou, Longpre, Zoph, Tay, Fedus, Li, Wang, Dehghani, Brahma et~al.}]{chung2022scaling}
Hyung~Won Chung, Le~Hou, Shayne Longpre, Barret Zoph, Yi~Tay, William Fedus, Yunxuan Li, Xuezhi Wang, Mostafa Dehghani, Siddhartha Brahma, et~al. 2022.
\newblock Scaling instruction-finetuned language models.
\newblock \emph{arXiv preprint arXiv:2210.11416}.

\bibitem[{Feng et~al.(2020)Feng, Chen, Lin, Wang, Yan, and Ren}]{feng2020scalable}
Yanlin Feng, Xinyue Chen, Bill~Yuchen Lin, Peifeng Wang, Jun Yan, and Xiang Ren. 2020.
\newblock Scalable multi-hop relational reasoning for knowledge-aware question answering.
\newblock \emph{arXiv preprint arXiv:2005.00646}.

\bibitem[{Hu et~al.(2021)Hu, Shen, Wallis, Allen-Zhu, Li, Wang, Wang, and Chen}]{hu2021lora}
Edward~J Hu, Yelong Shen, Phillip Wallis, Zeyuan Allen-Zhu, Yuanzhi Li, Shean Wang, Lu~Wang, and Weizhu Chen. 2021.
\newblock Lora: Low-rank adaptation of large language models.
\newblock \emph{arXiv preprint arXiv:2106.09685}.

\bibitem[{Huang et~al.(2022)Huang, Wu, Zhou, Gu, Zhao, and Cheng}]{huang2022clues}
Zixian Huang, Ao~Wu, Jiaying Zhou, Yu~Gu, Yue Zhao, and Gong Cheng. 2022.
\newblock Clues before answers: Generation-enhanced multiple-choice qa.
\newblock \emph{arXiv preprint arXiv:2205.00274}.

\bibitem[{Izacard et~al.(2022)Izacard, Lewis, Lomeli, Hosseini, Petroni, Schick, Dwivedi-Yu, Joulin, Riedel, and Grave}]{izacard2022few}
Gautier Izacard, Patrick Lewis, Maria Lomeli, Lucas Hosseini, Fabio Petroni, Timo Schick, Jane Dwivedi-Yu, Armand Joulin, Sebastian Riedel, and Edouard Grave. 2022.
\newblock Few-shot learning with retrieval augmented language models.
\newblock \emph{arXiv preprint arXiv:2208.03299}.

\bibitem[{Jiang et~al.(2023)Jiang, Xu, Gao, Sun, Liu, Dwivedi-Yu, Yang, Callan, and Neubig}]{jiang2023active}
Zhengbao Jiang, Frank~F Xu, Luyu Gao, Zhiqing Sun, Qian Liu, Jane Dwivedi-Yu, Yiming Yang, Jamie Callan, and Graham Neubig. 2023.
\newblock Active retrieval augmented generation.
\newblock \emph{arXiv preprint arXiv:2305.06983}.

\bibitem[{Jin et~al.(2021)Jin, Pan, Oufattole, Weng, Fang, and Szolovits}]{jin2021disease}
Di~Jin, Eileen Pan, Nassim Oufattole, Wei-Hung Weng, Hanyi Fang, and Peter Szolovits. 2021.
\newblock What disease does this patient have? a large-scale open domain question answering dataset from medical exams.
\newblock \emph{Applied Sciences}, 11(14):6421.

\bibitem[{Khashabi et~al.(2020)Khashabi, Min, Khot, Sabharwal, Tafjord, Clark, and Hajishirzi}]{khashabi2020unifiedqa}
Daniel Khashabi, Sewon Min, Tushar Khot, Ashish Sabharwal, Oyvind Tafjord, Peter Clark, and Hannaneh Hajishirzi. 2020.
\newblock Unifiedqa: Crossing format boundaries with a single qa system.
\newblock \emph{arXiv preprint arXiv:2005.00700}.

\bibitem[{Khattab et~al.(2022)Khattab, Santhanam, Li, Hall, Liang, Potts, and Zaharia}]{khattab2022demonstrate}
Omar Khattab, Keshav Santhanam, Xiang~Lisa Li, David Hall, Percy Liang, Christopher Potts, and Matei Zaharia. 2022.
\newblock Demonstrate-search-predict: Composing retrieval and language models for knowledge-intensive nlp.
\newblock \emph{arXiv preprint arXiv:2212.14024}.

\bibitem[{Lan et~al.(2019)Lan, Chen, Goodman, Gimpel, Sharma, and Soricut}]{lan2019albert}
Zhenzhong Lan, Mingda Chen, Sebastian Goodman, Kevin Gimpel, Piyush Sharma, and Radu Soricut. 2019.
\newblock Albert: A lite bert for self-supervised learning of language representations.
\newblock \emph{arXiv preprint arXiv:1909.11942}.

\bibitem[{Lewis et~al.(2020)Lewis, Perez, Piktus, Petroni, Karpukhin, Goyal, K{\"u}ttler, Lewis, Yih, Rockt{\"a}schel et~al.}]{lewis2020retrieval}
Patrick Lewis, Ethan Perez, Aleksandra Piktus, Fabio Petroni, Vladimir Karpukhin, Naman Goyal, Heinrich K{\"u}ttler, Mike Lewis, Wen-tau Yih, Tim Rockt{\"a}schel, et~al. 2020.
\newblock Retrieval-augmented generation for knowledge-intensive nlp tasks.
\newblock \emph{Advances in Neural Information Processing Systems}, 33:9459--9474.

\bibitem[{Li et~al.(2023)Li, Zhao, Chia, Ding, Bing, Joty, and Poria}]{li2023chain}
Xingxuan Li, Ruochen Zhao, Yew~Ken Chia, Bosheng Ding, Lidong Bing, Shafiq Joty, and Soujanya Poria. 2023.
\newblock Chain of knowledge: A framework for grounding large language models with structured knowledge bases.
\newblock \emph{arXiv preprint arXiv:2305.13269}.

\bibitem[{Li et~al.(2022)Li, Zhao, Lyu, and Wang}]{li2022eliciting}
Yanyang Li, Jianqiao Zhao, Michael~R Lyu, and Liwei Wang. 2022.
\newblock Eliciting knowledge from large pre-trained models for unsupervised knowledge-grounded conversation.
\newblock \emph{arXiv preprint arXiv:2211.01587}.

\bibitem[{Lin et~al.(2019)Lin, Chen, Chen, and Ren}]{lin2019kagnet}
Bill~Yuchen Lin, Xinyue Chen, Jamin Chen, and Xiang Ren. 2019.
\newblock Kagnet: Knowledge-aware graph networks for commonsense reasoning.
\newblock \emph{arXiv preprint arXiv:1909.02151}.

\bibitem[{Liu et~al.(2020)Liu, Shareghi, Meng, Basaldella, and Collier}]{liu2020self}
Fangyu Liu, Ehsan Shareghi, Zaiqiao Meng, Marco Basaldella, and Nigel Collier. 2020.
\newblock Self-alignment pretraining for biomedical entity representations.
\newblock \emph{arXiv preprint arXiv:2010.11784}.

\bibitem[{Liu et~al.(2021)Liu, Liu, Lu, Welleck, West, Bras, Choi, and Hajishirzi}]{liu2021generated}
Jiacheng Liu, Alisa Liu, Ximing Lu, Sean Welleck, Peter West, Ronan~Le Bras, Yejin Choi, and Hannaneh Hajishirzi. 2021.
\newblock Generated knowledge prompting for commonsense reasoning.
\newblock \emph{arXiv preprint arXiv:2110.08387}.

\bibitem[{Ma et~al.(2023)Ma, Gong, He, Zhao, and Duan}]{ma2023query}
Xinbei Ma, Yeyun Gong, Pengcheng He, Hai Zhao, and Nan Duan. 2023.
\newblock Query rewriting for retrieval-augmented large language models.
\newblock \emph{arXiv preprint arXiv:2305.14283}.

\bibitem[{Mihaylov et~al.(2018)Mihaylov, Clark, Khot, and Sabharwal}]{mihaylov2018can}
Todor Mihaylov, Peter Clark, Tushar Khot, and Ashish Sabharwal. 2018.
\newblock Can a suit of armor conduct electricity? a new dataset for open book question answering.
\newblock \emph{arXiv preprint arXiv:1809.02789}.

\bibitem[{Ouyang et~al.(2022)Ouyang, Wu, Jiang, Almeida, Wainwright, Mishkin, Zhang, Agarwal, Slama, Ray et~al.}]{ouyang2022training}
Long Ouyang, Jeffrey Wu, Xu~Jiang, Diogo Almeida, Carroll Wainwright, Pamela Mishkin, Chong Zhang, Sandhini Agarwal, Katarina Slama, Alex Ray, et~al. 2022.
\newblock Training language models to follow instructions with human feedback.
\newblock \emph{Advances in Neural Information Processing Systems}, 35:27730--27744.

\bibitem[{Peng et~al.(2023)Peng, Galley, He, Cheng, Xie, Hu, Huang, Liden, Yu, Chen et~al.}]{peng2023check}
Baolin Peng, Michel Galley, Pengcheng He, Hao Cheng, Yujia Xie, Yu~Hu, Qiuyuan Huang, Lars Liden, Zhou Yu, Weizhu Chen, et~al. 2023.
\newblock Check your facts and try again: Improving large language models with external knowledge and automated feedback.
\newblock \emph{arXiv preprint arXiv:2302.12813}.

\bibitem[{Petroni et~al.(2019)Petroni, Rockt{\"a}schel, Lewis, Bakhtin, Wu, Miller, and Riedel}]{petroni2019language}
Fabio Petroni, Tim Rockt{\"a}schel, Patrick Lewis, Anton Bakhtin, Yuxiang Wu, Alexander~H Miller, and Sebastian Riedel. 2019.
\newblock Language models as knowledge bases?
\newblock \emph{arXiv preprint arXiv:1909.01066}.

\bibitem[{Pirtoaca()}]{pirtoacaai2}
George~Sebastian Pirtoaca.
\newblock Ai2 leaderboard.
\newblock \emph{URL https://leaderboard. allenai. org/open\_book\_qa/submission/brhieieqaupc4cnddfg0}.

\bibitem[{Raffel et~al.(2020)Raffel, Shazeer, Roberts, Lee, Narang, Matena, Zhou, Li, and Liu}]{raffel2020exploring}
Colin Raffel, Noam Shazeer, Adam Roberts, Katherine Lee, Sharan Narang, Michael Matena, Yanqi Zhou, Wei Li, and Peter~J Liu. 2020.
\newblock Exploring the limits of transfer learning with a unified text-to-text transformer.
\newblock \emph{The Journal of Machine Learning Research}, 21(1):5485--5551.

\bibitem[{Robinson et~al.(2022)Robinson, Rytting, and Wingate}]{robinson2022leveraging}
Joshua Robinson, Christopher~Michael Rytting, and David Wingate. 2022.
\newblock Leveraging large language models for multiple choice question answering.
\newblock \emph{arXiv preprint arXiv:2210.12353}.

\bibitem[{Santoro et~al.(2017)Santoro, Raposo, Barrett, Malinowski, Pascanu, Battaglia, and Lillicrap}]{santoro2017simple}
Adam Santoro, David Raposo, David~G Barrett, Mateusz Malinowski, Razvan Pascanu, Peter Battaglia, and Timothy Lillicrap. 2017.
\newblock A simple neural network module for relational reasoning.
\newblock \emph{Advances in neural information processing systems}, 30.

\bibitem[{Schlichtkrull et~al.(2018)Schlichtkrull, Kipf, Bloem, Van Den~Berg, Titov, and Welling}]{schlichtkrull2018modeling}
Michael Schlichtkrull, Thomas~N Kipf, Peter Bloem, Rianne Van Den~Berg, Ivan Titov, and Max Welling. 2018.
\newblock Modeling relational data with graph convolutional networks.
\newblock In \emph{The Semantic Web: 15th International Conference, ESWC 2018, Heraklion, Crete, Greece, June 3--7, 2018, Proceedings 15}, pages 593--607. Springer.

\bibitem[{Shwartz et~al.(2020)Shwartz, West, Bras, Bhagavatula, and Choi}]{shwartz2020unsupervised}
Vered Shwartz, Peter West, Ronan~Le Bras, Chandra Bhagavatula, and Yejin Choi. 2020.
\newblock Unsupervised commonsense question answering with self-talk.
\newblock \emph{arXiv preprint arXiv:2004.05483}.

\bibitem[{Speer et~al.(2017)Speer, Chin, and Havasi}]{speer2017conceptnet}
Robyn Speer, Joshua Chin, and Catherine Havasi. 2017.
\newblock Conceptnet 5.5: An open multilingual graph of general knowledge.
\newblock In \emph{Proceedings of the AAAI conference on artificial intelligence}, volume~31.

\bibitem[{Talmor et~al.(2018)Talmor, Herzig, Lourie, and Berant}]{talmor2018commonsenseqa}
Alon Talmor, Jonathan Herzig, Nicholas Lourie, and Jonathan Berant. 2018.
\newblock Commonsenseqa: A question answering challenge targeting commonsense knowledge.
\newblock \emph{arXiv preprint arXiv:1811.00937}.

\bibitem[{Taunk et~al.(2023)Taunk, Khanna, Kandru, Varma, Sharma, and Tapaswi}]{taunk2023grapeqa}
Dhaval Taunk, Lakshya Khanna, Siri Venkata Pavan~Kumar Kandru, Vasudeva Varma, Charu Sharma, and Makarand Tapaswi. 2023.
\newblock Grapeqa: Graph augmentation and pruning to enhance question-answering.
\newblock In \emph{Companion Proceedings of the ACM Web Conference 2023}, pages 1138--1144.

\bibitem[{Touvron et~al.(2023)Touvron, Martin, Stone, Albert, Almahairi, Babaei, Bashlykov, Batra, Bhargava, Bhosale et~al.}]{touvron2023llama}
Hugo Touvron, Louis Martin, Kevin Stone, Peter Albert, Amjad Almahairi, Yasmine Babaei, Nikolay Bashlykov, Soumya Batra, Prajjwal Bhargava, Shruti Bhosale, et~al. 2023.
\newblock Llama 2: Open foundation and fine-tuned chat models.
\newblock \emph{arXiv preprint arXiv:2307.09288}.

\bibitem[{Wang et~al.(2022)Wang, Chan, Ilievski, Chen, and Ren}]{wang2022pinto}
Peifeng Wang, Aaron Chan, Filip Ilievski, Muhao Chen, and Xiang Ren. 2022.
\newblock Pinto: Faithful language reasoning using prompt-generated rationales.
\newblock \emph{arXiv preprint arXiv:2211.01562}.

\bibitem[{Wang et~al.(2020)Wang, Peng, Ilievski, Szekely, and Ren}]{wang2020connecting}
Peifeng Wang, Nanyun Peng, Filip Ilievski, Pedro Szekely, and Xiang Ren. 2020.
\newblock Connecting the dots: A knowledgeable path generator for commonsense question answering.
\newblock \emph{arXiv preprint arXiv:2005.00691}.

\bibitem[{Wang et~al.(2019)Wang, Kapanipathi, Musa, Yu, Talamadupula, Abdelaziz, Chang, Fokoue, Makni, Mattei et~al.}]{wang2019improving}
Xiaoyan Wang, Pavan Kapanipathi, Ryan Musa, Mo~Yu, Kartik Talamadupula, Ibrahim Abdelaziz, Maria Chang, Achille Fokoue, Bassem Makni, Nicholas Mattei, et~al. 2019.
\newblock Improving natural language inference using external knowledge in the science questions domain.
\newblock In \emph{Proceedings of the AAAI Conference on Artificial Intelligence}, volume~33, pages 7208--7215.

\bibitem[{West et~al.(2021)West, Bhagavatula, Hessel, Hwang, Jiang, Bras, Lu, Welleck, and Choi}]{west2021symbolic}
Peter West, Chandra Bhagavatula, Jack Hessel, Jena~D Hwang, Liwei Jiang, Ronan~Le Bras, Ximing Lu, Sean Welleck, and Yejin Choi. 2021.
\newblock Symbolic knowledge distillation: from general language models to commonsense models.
\newblock \emph{arXiv preprint arXiv:2110.07178}.

\bibitem[{Wishart et~al.(2018)Wishart, Feunang, Guo, Lo, Marcu, Grant, Sajed, Johnson, Li, Sayeeda et~al.}]{wishart2018drugbank}
David~S Wishart, Yannick~D Feunang, An~C Guo, Elvis~J Lo, Ana Marcu, Jason~R Grant, Tanvir Sajed, Daniel Johnson, Carin Li, Zinat Sayeeda, et~al. 2018.
\newblock Drugbank 5.0: a major update to the drugbank database for 2018.
\newblock \emph{Nucleic acids research}, 46(D1):D1074--D1082.

\bibitem[{Xu et~al.(2021{\natexlab{a}})Xu, Zhang, Cai, and Lam}]{xu2021dynamic}
Weiwen Xu, Huihui Zhang, Deng Cai, and Wai Lam. 2021{\natexlab{a}}.
\newblock Dynamic semantic graph construction and reasoning for explainable multi-hop science question answering.
\newblock \emph{arXiv preprint arXiv:2105.11776}.

\bibitem[{Xu et~al.(2021{\natexlab{b}})Xu, Zhu, Wang, Sun, Cheng, Liu, Gao, He, Zeng, and Huang}]{xu2021human}
Yichong Xu, Chenguang Zhu, Shuohang Wang, Siqi Sun, Hao Cheng, Xiaodong Liu, Jianfeng Gao, Pengcheng He, Michael Zeng, and Xuedong Huang. 2021{\natexlab{b}}.
\newblock Human parity on commonsenseqa: Augmenting self-attention with external attention.
\newblock \emph{arXiv preprint arXiv:2112.03254}.

\bibitem[{Xu et~al.(2021{\natexlab{c}})Xu, Zhu, Xu, Liu, Zeng, and Huang}]{xu2021fusing}
Yichong Xu, Chenguang Zhu, Ruochen Xu, Yang Liu, Michael Zeng, and Xuedong Huang. 2021{\natexlab{c}}.
\newblock Fusing context into knowledge graph for commonsense question answering.
\newblock In \emph{Findings of the Association for Computational Linguistics: ACL-IJCNLP 2021}, pages 1201--1207.

\bibitem[{Yan et~al.(2020)Yan, Raman, Chan, Zhang, Rossi, Zhao, Kim, Lipka, and Ren}]{yan2020learning}
Jun Yan, Mrigank Raman, Aaron Chan, Tianyu Zhang, Ryan Rossi, Handong Zhao, Sungchul Kim, Nedim Lipka, and Xiang Ren. 2020.
\newblock Learning contextualized knowledge structures for commonsense reasoning.
\newblock \emph{arXiv preprint arXiv:2010.12873}.

\bibitem[{Yao et~al.(2022)Yao, Zhao, Yu, Du, Shafran, Narasimhan, and Cao}]{yao2022react}
Shunyu Yao, Jeffrey Zhao, Dian Yu, Nan Du, Izhak Shafran, Karthik Narasimhan, and Yuan Cao. 2022.
\newblock React: Synergizing reasoning and acting in language models.
\newblock \emph{arXiv preprint arXiv:2210.03629}.

\bibitem[{Yao et~al.(2023)Yao, Wang, Mao, Tan, Huang, Chen, and Zhang}]{yao2023knowledge}
Yunzhi Yao, Peng Wang, Shengyu Mao, Chuanqi Tan, Fei Huang, Huajun Chen, and Ningyu Zhang. 2023.
\newblock Knowledge rumination for pre-trained language models.
\newblock \emph{arXiv preprint arXiv:2305.08732}.

\bibitem[{Yasunaga et~al.(2022)Yasunaga, Bosselut, Ren, Zhang, Manning, Liang, and Leskovec}]{yasunaga2022deep}
Michihiro Yasunaga, Antoine Bosselut, Hongyu Ren, Xikun Zhang, Christopher~D Manning, Percy~S Liang, and Jure Leskovec. 2022.
\newblock Deep bidirectional language-knowledge graph pretraining.
\newblock \emph{Advances in Neural Information Processing Systems}, 35:37309--37323.

\bibitem[{Yasunaga et~al.(2021)Yasunaga, Ren, Bosselut, Liang, and Leskovec}]{yasunaga2021qa}
Michihiro Yasunaga, Hongyu Ren, Antoine Bosselut, Percy Liang, and Jure Leskovec. 2021.
\newblock Qa-gnn: Reasoning with language models and knowledge graphs for question answering.
\newblock \emph{arXiv preprint arXiv:2104.06378}.

\bibitem[{Yu et~al.(2023)Yu, Zhang, Liang, Jiang, and Sabharwal}]{yu2023improving}
Wenhao Yu, Zhihan Zhang, Zhenwen Liang, Meng Jiang, and Ashish Sabharwal. 2023.
\newblock Improving language models via plug-and-play retrieval feedback.
\newblock \emph{arXiv preprint arXiv:2305.14002}.

\bibitem[{Zelikman et~al.(2022)Zelikman, Wu, Mu, and Goodman}]{zelikman2022star}
Eric Zelikman, Yuhuai Wu, Jesse Mu, and Noah Goodman. 2022.
\newblock Star: Bootstrapping reasoning with reasoning.
\newblock \emph{Advances in Neural Information Processing Systems}, 35:15476--15488.

\bibitem[{Zhang et~al.(2023)Zhang, Pan, Zhao, and Wang}]{zhang2023mitigating}
Shuo Zhang, Liangming Pan, Junzhou Zhao, and William~Yang Wang. 2023.
\newblock Mitigating language model hallucination with interactive question-knowledge alignment.
\newblock \emph{arXiv preprint arXiv:2305.13669}.

\bibitem[{Zhang et~al.(2022)Zhang, Bosselut, Yasunaga, Ren, Liang, Manning, and Leskovec}]{zhang2022greaselm}
Xikun Zhang, Antoine Bosselut, Michihiro Yasunaga, Hongyu Ren, Percy Liang, Christopher~D Manning, and Jure Leskovec. 2022.
\newblock Greaselm: Graph reasoning enhanced language models for question answering.
\newblock \emph{arXiv preprint arXiv:2201.08860}.

\end{thebibliography}
\bibliographystyle{acl_natbib}

\appendix

\section{Appendix}
\label{sec:appendix}
\begin{table*}[h]
\centering
\begin{tabular}{|c|c|}
\hline Dataset & Example \\
\hline CommonsenseQA & \begin{tabular}{l} 
A weasel has a thin body and short legs to easier burrow after prey in a what? \\
(A) tree (B) mulberry bush (C) chicken coop (D) viking ship \textcolor[rgb]{1,0,0}{(E) rabbit warren}
\end{tabular} \\
\hline OpenbookQA & \begin{tabular}{l} 
Which of these would let the most heat travel through? \\
\begin{tabular}{ll} 
(A) a new pair of jeans & \textcolor[rgb]{1,0,0}{(B) a steel spoon in a cafeteria} \\
(C) a cotton candy at a store & (D) a calvin klein cotton hat
\end{tabular}
\end{tabular} \\
\hline MedQA-USMLE & \begin{tabular}{l} 
A 57 -year-old man presents to his primary care physician with a 2-month \\
history of right upper and lower extremity weakness. He noticed the weakness \\
when he started falling far more frequently while running errands. Since then, \\
he has had increasing difficulty with walking and lifting objects. His past \\
medical history is significant only for well-controlled hypertension, but he says \\
that some members of his family have had musculoskeletal problems. His right \\
upper extremity shows forearm atrophy and depressed reflexes while his right \\
lower extremity is hypertonic with a positive Babinski sign. Which of the \\
following is most likely associated with the cause of this patients symptoms? \\
\begin{tabular}{ll} 
(A) HLA-B8 haplotype & (B) HLA-DR2 haplotype \\
\textcolor[rgb]{1,0,0}{(C) Mutation in SOD1} & (D) Mutation in SMN1
\end{tabular}
\end{tabular} \\
\hline
\end{tabular}
\caption{Examples of the Knowledge-intensive QA task for each of the datasets evaluated in this work. }
\end{table*}

\end{document}